\documentclass{llncs}
\usepackage[utf8]{inputenc}
\usepackage{hyperref}
\usepackage{url}
\usepackage{amssymb}
\usepackage{amsmath}
\usepackage{algorithm}
\usepackage[noend]{algpseudocode} 

\usepackage{color}
\usepackage{multirow}

\usepackage{xcolor}

\usepackage{graphicx}
\usepackage{subfig}
\usepackage{booktabs}

\begin{document}

\title{Towards Representation Learning with Tractable Probabilistic Models}
\titlerunning{Towards Representation Learning with Tractable Probabilistic Models}  

\author{Antonio Vergari \and Nicola Di Mauro \and Floriana Esposito}
\authorrunning{Antonio Vergari et al.} 

\institute{Department of Computer Science, University of Bari, Italy\\
  \email{\{antonio.vergari, nicola.dimauro, floriana.esposito\}@uniba.it}}

\maketitle 

\begin{abstract}
  Probabilistic models learned as density estimators can be exploited
  in representation learning beside being toolboxes
  used to answer inference queries only.
  However, how to extract useful representations highly depends on the
  particular model involved.
  We argue that tractable inference, i.e. inference that can be
  computed in polynomial time, can enable \emph{general} schemes to extract
  features from black box models.
  We plan to investigate how Tractable Probabilistic Models (TPMs) can be
  exploited to generate embeddings by random query evaluations.
  We devise two experimental designs to assess and compare different TPMs as
  feature extractors in an unsupervised representation learning
  framework.
  We show some experimental results on standard image datasets by
  applying such a method to Sum-Product Networks and Mixture of Trees
  as tractable models generating embeddings.
\end{abstract}

\section{Background And Motivation}
Density estimation is the unsupervised task of learning a model $\theta$ estimating a
joint probability distribution $p$ over a set of random variables
(r.v.s) $\mathbf{X}$ that are assumed to have generated a given set of observed
samples $\{\mathbf{x}^{i}\}_{i=1}^{m}$.
Once such an estimator is learned, one can use it to do
\emph{inference}, that is computing  the probability of queries about
certain states of the r.v.s,
e.g. the marginals $p_{\theta}(\mathbf{Q}), \mathbf{Q}\subseteq\mathbf{X}$.
Many machine learning problems can be reframed as different kinds of
inference tasks;
for instance, the classification of a set of r.v.s $\mathbf{Y}$ can be done
by Most Probable Explanation (MPE)
inference~\cite{Koller2009}.

Examples of commonly used density estimators comprise Probabilistic
Graphical Models (PGMs)~\cite{Koller2009} like Bayesian Networks (BNs)
and Markov Networks (MNs), that represent the conditional dependence
assumptions in $p$ in a graph formalism.
On the other hand, (deep) neural models represent a factorization of $p$
by the means of neural architectures like Restricted Boltzmann
Machines (RBMs) and its variations~\cite{Marlin2010}, Fully Visible Sigmoid Belief
Networks~\cite{Frey1998,Bengio2000} or embed an inference
algorithm into a network evaluation, like in the case of Variational Autoencoders~\cite{Kingma2013}.
Accurately estimating \emph{complex} distributions from data
and \emph{efficiently} querying them are the required qualities for a good
density estimator.
Performing exact inference is, however, still a hard task in
general. For PGMs, for example, even approximate inference routines
may end up being exponential in the worst case~\cite{Roth1996}. 

We want to investigate how to extract useful representations
from general purpose density estimators.
That is, finding a way to exploit probabilistic models, already
learned to encode $p$ in an unsupervised fashion by using existing
learning algorithms, in order to transform the initial data into another
representation space.
More specifically, we want to build an \emph{embedding} for each sample,
$\mathbf{e}^{i}=f_{\theta, p}(\mathbf{x}^{i})$,
such that the transformation $f_{}$ is provided by a probabilistic model
$\theta$, learned to estimate $p$.
These embeddings could be employed in other tasks such as
clustering and classification, allowing for new transfer and
 unsupervised learning schemas, as usually done in representation
 learning~\cite{Bengio2012}.
 The usefulness of these representations can be stated by proxy
 performance metrics on the subsequent supervised tasks, e.g. the
 accuracy scored on predicting some previously unseen sample
 labels~\cite{Bengio2012}.
 Moreover, they could enable new ways to assess and compare different models,
  going beyond the classic likelihood comparison, which can be
 highly misleading sometimes~\cite{Theis2016}.
 
To build a mapping $f_{\theta, p}$, one can take advantage of the
geometric space induced by $p_{\theta}$, the estimate of $p$ according to model $\theta$.
A straightforward way is to build each embedding component as the
result of a single
inference step on $\theta$ according to some query
$\mathbf{Q}=\mathbf{q}_j$, i.e. $e^{i}_{j}=p_{\theta}(\mathbf{q}_j)$.
The classic way in which this has been implemented in the past is to
employ the probability space embeddings in the construction of
hand-crafted and problem dependent kernels, like P-kernels~\cite{Shawe-Taylor2004}. 
Another possibility is to leverage additional first order information as
the one encoded in the
gradient of $p$, as already done by Fisher vectors and Fisher kernels~\cite{Jaakkola1999}.
However, to implement these classic approaches, one has to derive analytically the
embedding computation according to each
model parameters, a task not always feasible for all
models~\cite{Shawe-Taylor2004}.

We are interested in a generic procedure that employs different
density estimators as black boxes, despite their parametrizations or
inner representations.  
By following the first line of thought, we try to capture the differences
among the samples projected in various spaces induced by the
probability densities encoded in a model $\theta$.
The basic idea is to construct features with a random query generator,
computing their values according to the evaluation of $\theta$.
Hence, \emph{tractable} inference routines become crucial to construct complex
enough representations by evaluating $\theta$ several times.

In the following sections we introduce Tractable Probabilistic Models
and we define two possible schemas to generate embeddings from them by
leveraging the tractability property.
We then discuss a possible empirical evaluation of such a framework
and show some results for an experimental application on benchmark
image datasets, showing promising results.

\section{Tractable Probabilistic Models}

Tractable Probabilistic Models (TPMs) are density estimators for which
exact inference is polynomial in $|\mathbf{X}|$.
It is important to note that tractability of inference is not a global property, but it is associated
to classes of queries. For instance, pointwise evidence, marginals,
MPE inference and the computation of a partition function, etc can
each result
tractable for some models while the others are unfeasible for the same
models~\cite{Darwiche2009}.

Here we provide a rough classification and review for some TPMs commonly found in
the literature: 
\begin{itemize}
\item \emph{low treewidth} PGMs, alleviating the computation of the
  partition function of $p$ by limiting the model expressiveness in
  terms of the representable conditional dependencies.
  They comprise models such as Hidden Markov Models~\cite{Koller2009}, tree distributions
  and their mixture~\cite{Meila2000} or latent r.v. variants~\cite{Choi2011},
  for which pointwise and marginal queries can be answered in time linear to the number of
  r.v.s.  
\item computational models derived from a \emph{knowledge compilation} process, involving
   the elicitation of another representation form for $p$ (e.g. the
   network polynomial), usually
   encoding it into a
   computation graph. This is the case of Arithmetic Circuits (ACs)~\cite{Darwiche2009},
   and Sentential Decision Diagrams (SDDs)~\cite{Darwiche2011},
   both enabling intractable BNs and MNs to be compiled into data
   structures for which marginals and even MPE inference can be
   answered in time linear in the size of such structures~\cite{Lowd2013}.
\item \emph{neural autoregressive} models, factorizing $p$ according
  to the chain rule and modeling each factor by a (possibly deep)
  neural network. Models like Neural Autoregressive Distribution
  Estimators (NADEs)~\cite{Larochelle2011} and
  Masked Autoencoder Distribution Estimators (MADE)~\cite{Germain2015} leverage constrained feedfoward and
  autoencoder networks to provide pointwise inference linear in the
  size of the networks.
  To answer marginal queries in polynomial time, variants like EoNADE, allowing
  for an order agnostic training of the factors, are necessary~\cite{Uria2013}.
\end{itemize}
Complete evidences, marginals, and even the computation of the partition
function for $p$ are computable in tractable time for Sum-Product Networks
(SPNs)~\cite{Poon2011a}.
SPNs are deep neural architectures equivalent to ACs for finite domains,
compiling the network polynomial of $p$ into a more sophisticated
architecture that introduces a latent variable semantics~\cite{Peharz2015a,Peharz2016}.
Differently from other tractable neural models, SPNs guarantee that
each hidden neuron still models a correct probability distribution
over a restriction of the input r.v.s $\mathbf{X}$, named its \emph{scope}.
This constrained form allows for a \emph{direct encoding} of the input
space.
The great interest around SPNs is also motivated by the increasing
arsenal of structure learning algorithms arising in the
literature~\cite{Gens2013,Vergari2015}.
This makes SPNs one of the few deep architectures for which the
structure can be directly and effectively learned and not crafted or
tuned by hand.

\section{Representation Learning for TPMs}

For tractable models like HMMs one can leverage the emission
probabilities for each evaluated sample to generate a
particular kernel~\cite{Shawe-Taylor2004}. 
This case is an example of a model dependent embedding
space construction, in which not only the tractability of the model is
mandatory, but also the input space definition shall be suitable
(e.g. sequences as samples). In the same way, building a Fisher kernel
for Gaussian Mixture Models can be done easily because it is possible
to compute the gradient of the likelihood in a closed form for each
parameter.

For neural architectures, representations are usually extracted as the
activations of hidden neurons after the evaluation of each input
sample.
It is common practice to extract the embeddings from their last layers,
comprising higher level representations, hence the more informative
ones~\cite{Bengio2012}.
While one could adopt the same approach for
other neural autoregressive density estimators, the probabilistic
meaning of the extracted features would be lost and a direct
comparison against other models would not be possible.
Extracting node activations from an SPN would retain their probabilistic
interpretation, however they would highly depend on the (structure or
weight) learning algorithm employed to train the model.
Moreover, it is still not clear which hidden neurons to choose
to extract meaningful embeddings from an SPN.
Following a layer-wise procedure for the embedding extraction does not cope with the
different scope semantics of an SPN.
As a general rule of thumb, one could employ all the nodes or all the
nodes of one kind (sum or products only).

We argue that different TPMs as black box density estimators can be used to
generate vector embeddings on a common ground, by defining a set of
template queries to be answered.
By doing so, one is able to evaluate different regions of the
probability density surfaces induced by different queries.
For instance, defining marginalized conditional queries would represent probability
distributions different from the joint $p$, highlighting local
interactions for the input r.v.s.

The most basic way to do so is to generate these queries in a random
fashion.
However, defining more sophisticated schemes is possible.
In the end, a domain knowledge guided embedding extraction process
could be thought as a sort of feature engineering step in which query template
families are devised as features.

We propose two different embedding generation schemas exploiting randomness.
In the first one, each feature template is constructed as a random marginal
  query evaluation, i.e. a subset of r.v.s
  $\mathbf{Q}_{j}\subseteq\mathbf{X}, j=\,\dots,k$ is randomly chosen for each 
  component of the $k$-dimensional embedding to generate.
  Each sample embedding is then exactly computed as
  $e^{i}_{j}=p_{\theta}(\mathbf{Q}_{j}=\mathbf{x}^{i}_{\mathbf{Q}_{j}})$,
  where $\mathbf{x}^{i}_{\mathbf{Q}_{j}}$ indicates the restriction of
  the sample vector $\mathbf{x}^{i}$ to the r.v.s in $\mathbf{Q}_{j}$.
  Again, the random selection criterion can be lead by domain
  knowledge, e.g. if the samples $\mathbf{x}^{i}$ are images, it is
  meaningful to extract adjacent pixels as the r.v.s $\mathbf{Q}_{j}$.
  A sketch of this process is presented in Algorithm~\ref{algo:randQuery}.

  The second approach, presented as Algorithm~\ref{algo:randPatch}, extracts a set of random ``patches''
  from the training data, as $s$ random portions of size $d$,
  $\{\mathbf{r}^{i}\}_{i=1}^{s}$, from randomly chosen input samples.
  Then a TPM $\theta$ is trained on this reduced dataset.
  Each embedding component is then generated by
  evaluating the original samples for different subsets
  $\{\mathbf{Q}_{h}\subset \mathbf{X}\}_{h=1}^{w}, |\mathbf{Q}_{h}|=d$
  of the training samples of length $d$, according to model $\theta$.
  For instance, in a sliding window approach, with unitary stride,
  the total number of adjacent subset of length $d$ from a vector of size $n$ would be $w=n-d$.
  This is more meaningful for image samples as it is a way to take
  into account pixel autocorrelation and translation invariance.
  If that were the case, this approach would be similar to the
  dictionary learning one proposed in~\cite{Coates2011}.
  In this case, however, the feature generation scheme is directly done by a pointwise
  evaluation of a whole TPM, which in turn is directly learned on the patches.

\begin{algorithm}[!t]
  \caption{\textsf{randQueryEmbedding}($\mathcal{D}$, $k$)}
  \label{algo:randQuery}
  \begin{algorithmic}[1]
    \State \textbf{Input:} a set of instances
    $\mathcal{D}=\{\mathbf{x}^{i}\}_{i=1}^{m}$ over r.v.s
    $\mathbf{X}=\{X_1,\dots,X_n\}$, $k$ as the number of features to generate
  \State  \textbf{Output:}  a set of embeddings
  $\mathcal{E}=\{\mathbf{e}^{i}\}_{i=1}^{m}, \mathbf{e}^{i}\in\mathbb{R}^{k}$

  \State $\theta\leftarrow\mathsf{learnDensityEstimator}(\mathcal{D})$
  \State $\mathcal{E}\leftarrow\{\}$
  \For {$j=1,\dots,k$}
  \State $\mathbf{Q}_{j}\leftarrow \mathsf{selectRandomRVs}(\mathbf{X})$
  \For {$i=1,\dots,m$}
  \State $e^{i}_{j}= p_{\theta}(\mathbf{x}^{i}_{\mathbf{Q}_{j}})$  
  \EndFor
  \State $\mathcal{E}\leftarrow\mathcal{E}\cup\{\mathbf{e}^{i}\}$
  \EndFor
  \Return {$\mathcal{E}$}
      \end{algorithmic}
\end{algorithm}

\begin{algorithm}[!t]
  \caption{\textsf{randPatchEmbedding}($\mathcal{D}$, $s$, $d$)}
  \label{algo:randPatch}
  \begin{algorithmic}[1]
    \State \textbf{Input:} a set of instances
    $\mathcal{D}=\{\mathbf{x}^{i}\}_{i=1}^{m}$ over r.v.s
    $\mathbf{X}=\{X_1,\dots,X_n\}$,
    $s$ as the number of patches to extract,
    $d$ as the patch length,
    \State  \textbf{Output:}  a set of embeddings
    $\mathcal{E}=\{\mathbf{e}^{i}\}_{i=1}^{m}, \mathbf{e}^{i}\in\mathbb{R}^{k}$
    \State $\mathcal{R}\leftarrow\{\}$
    \For {$i=1,\dots,s$}
    \State $\mathbf{x}^{\mathsf{rand}}\leftarrow
    \mathsf{selectRandomSample}(\mathcal{D})$
    \State $\mathbf{r}^{i}\leftarrow
    \mathsf{extractRandomPatch}(\mathbf{x}^{\mathsf{rand}}, d)$
    \State $\mathcal{R}\leftarrow\mathcal{R}\cup\{\mathbf{r}^{i}\}$
    \EndFor
    
    \State $\theta\leftarrow\mathsf{learnDensityEstimator}(\mathcal{R})$

    \State $\mathcal{E}\leftarrow\{\}$
    \For {$i=1,\dots,m$}
    \State $j\leftarrow 0$
    \For {\textbf{each} patch $\mathbf{q}^{i}, |\mathbf{q}^{i}|=d$ in
      $\mathbf{x}^{i}$}
    \State $e^{i}_{j}= p_{\theta}(\mathbf{q}^{i})$
    \State $j\leftarrow j + 1$
    \EndFor
    \State $\mathcal{E}\leftarrow\mathcal{E}\cup\{\mathbf{e}^{i}\}$
    \EndFor
    \Return {$\mathcal{E}$}
  \end{algorithmic}
\end{algorithm}

In the first approach it is mandatory that the models involved
guarantee tractable marginal inference.
Indeed, in a naive computation one would have to ask a model
$k\cdot m$ marginal queries, and even if one cashes the answers for
each possible observable state configuration for the r.v.s $\mathbf{Q}_{j}$,
assuming all of them to have at most $l$ values, $l < m$, it would end
up with $k\cdot l$ evaluations.
In a random generation approach, the value of $k$ will probably be
large enough to demand each single evaluation to be as fast as possible.

The first approach is also the more flexible one,
since it can be adapted to other types of queries, possibly involving
more complex computations but still tractable for certain models.
For example, symmetric queries like parity counting, end up being
tractable for SDDs~\cite{Bekker2015}.

On the other hand, the second one requires only tractable pointwise
inference.
Therefore, it is more easily adaptable to a larger set of models,
since all locally learned models are globally evaluated and no
marginal computations are involved.

Both models allow for very different TPMs to be evaluated against the
same exact set of feature templates, since the generation of the
random queries and patches can be done just once in a comparative
experiment.
This aspect allows for fairer comparisons among different models, moreover,
it enables an additional way to understand the structure of such
models: feature selection routines applied on the generated embedding
spaces could reveal particularly effective sets of features hinting at
powerful r.v.s interactions.

\section{Evaluation}

We plan to conduct a thorough empirical evaluation comprising at least
two TPMs for each of the kinds presented in the second section.
The meaningfulness and usefulness of the generated embeddings shall be
measured against different metrics and in different tasks.
For instance, both multi-class and multi-label classification tasks
are worth investigating, and for each one of them, several metrics
like accuracy, hamming loss and exact match shall be taken into account.
The main objective of this intensive empirical comparison is to
investigate how different TPMs would behave under this general
embedding extraction framework.

Here we present a smaller empirical evaluation of the first proposed approach
in a classification task on three standard benchmark image datasets.
We employ both SPNs and Mixture of Trees (MT)~\cite{Meila2000} as a PGM
for which marginal inference is tractable.
We want to determine how different models, with different expressive
capabilities, generate embeddings against the \emph{same} marginal query
evaluations.
To learn a supervised classifier, we employ a logistic
regressor  with an L2 regularizer in a one-versus-rest setting on top
of all the embeddings we generate.
We use the implementation available in the
\textsf{scikit-learn}
framework\footnote{\url{http://scikit-learn.org/}}.
For each representation we determine the regularization
coefficient $C$ value in $\{0.0001, 0.001, 0.01, 0.1,
1.0\}$, choosing the model with the best validation accuracy.
As the simplest baseline possible, we apply such a classifier
directly on the initial data representation $\mathbf{x}^{i}$, denoting
it with \textsf{LR}.
To declare an extracted representation as useful, we expect it to
surpass the accuracy score of \textsf{LR}.

We consider the following datasets:
Rectangles (\textsf{REC}) as introduced in~\cite{Larochelle2007},
Convex (\textsf{CON}), always introduced in~\cite{Larochelle2007},
OCR Letters (\textsf{OCR}) as
presented in~\cite{Larochelle2010},
Caltech-101 Silhouettes (\textsf{CAL})~\cite{Marlin2010} and a binarized version
of MNIST (\textsf{BMN}) as processed in~\cite{Salakhutdinov2008}.
The accuracy scores for the best \textsf{LR} models are 75.58\% for
\textsf{OCR}, 62.67\% for \textsf{CAL} and 90.62\% for \textsf{BMN}.

We employ models of the same family but with different model capacities
to study the effect of regularization on the same set of features.
We learn three differently regularized SPN models on each dataset, by
employing \textsf{LearnSPN-b}~\cite{Vergari2015}, a structure learning
algorithm adopting a simplified iterated clustering strategy for
determine the insertion of the inner nodes in the networks.
We let vary the early stopping parameter $m\in\{500, 100, 50\}$ while we
fix the G-test threshold $\rho$ to 15 for \textsf{OCR}
and 20 for all the other datasets.
The effect of parameter $m$ is to regularize the learned SPNs,
therefore it governs the model capacity.
We denote these models as \textsf{SPN-I}, \textsf{SPN-II}
and \textsf{SPN-III} respectively.
For MT, we let the number of the mixture component vary among 3, 15
and 30, ending up with three models, \textsf{MT-I}, \textsf{MT-II},
\textsf{MT-III}.

We evaluate all the models on 1000 randomly generated marginal queries on each dataset.
For each query we select r.v.s corresponding to
adjacent pixels in a rectangular image patch having minimum sizes of 2
pixels and maximum of 7 pixels for \textsf{OCR} and 10 pixels for the
remaining datasets.

\begin{figure}[!t]
  \centering
  \subfloat[]{\includegraphics[width=0.33\columnwidth]
    {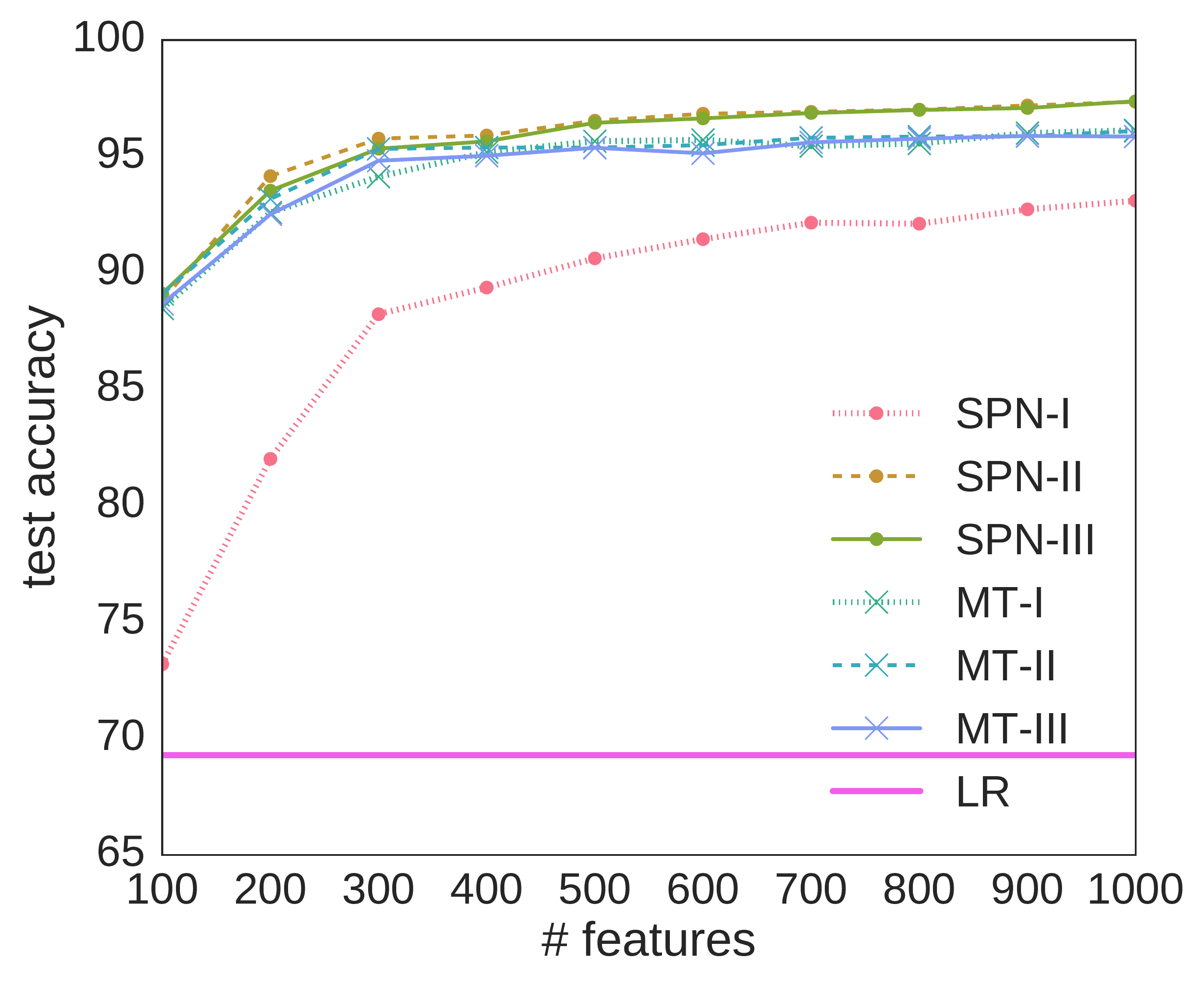}
    \label{fig:rect-lines-rec}}\hspace{-10pt}
  \subfloat[]{\includegraphics[width=0.33\columnwidth]
    {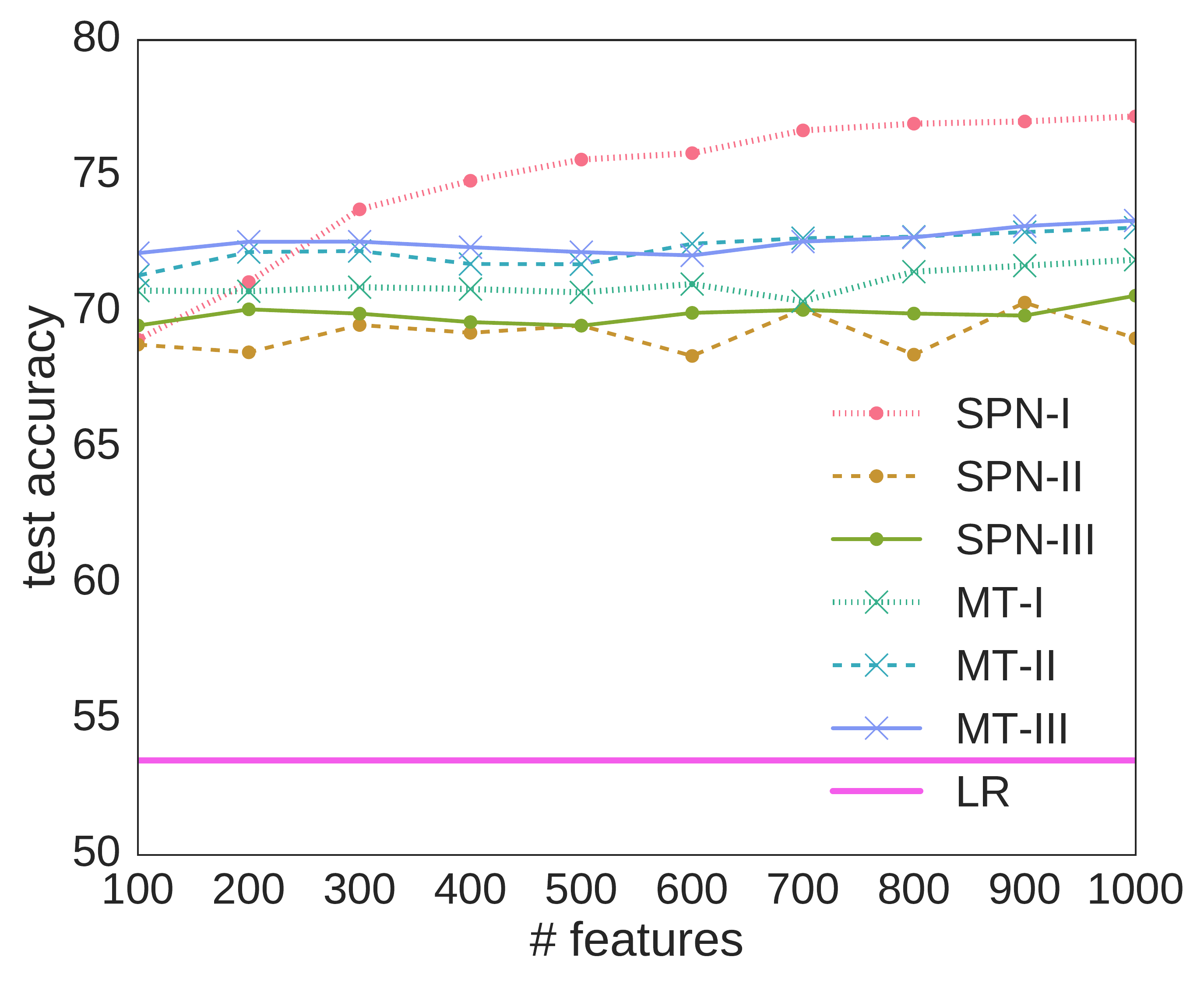}
    \label{fig:rect-lines-con}}\hspace{-10pt}
  \subfloat[]{\includegraphics[width=0.33\columnwidth]
    {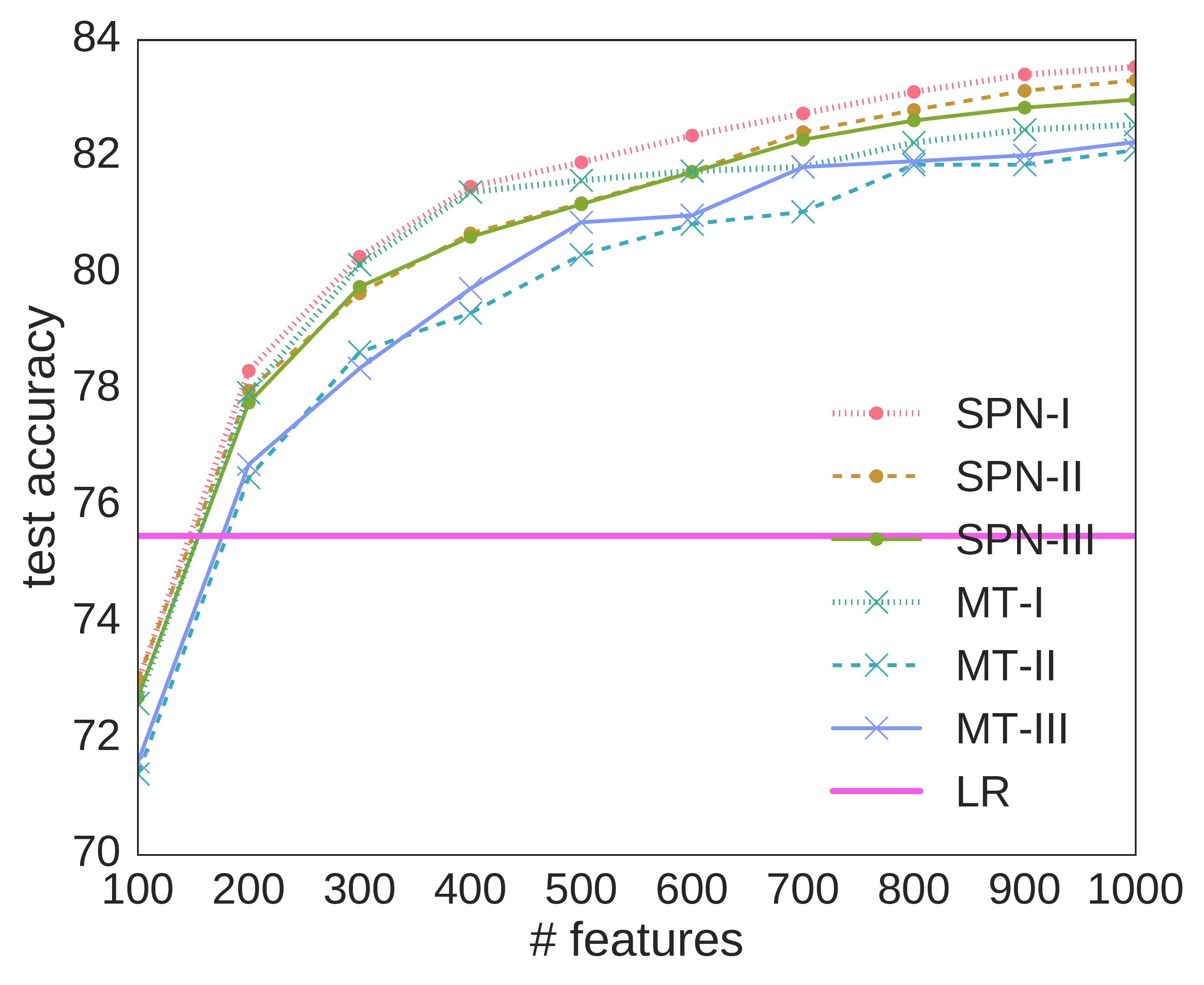}
    \label{fig:rect-lines-ocr}}\\[-5pt]
  \subfloat[]{\includegraphics[width=0.33\columnwidth]
    {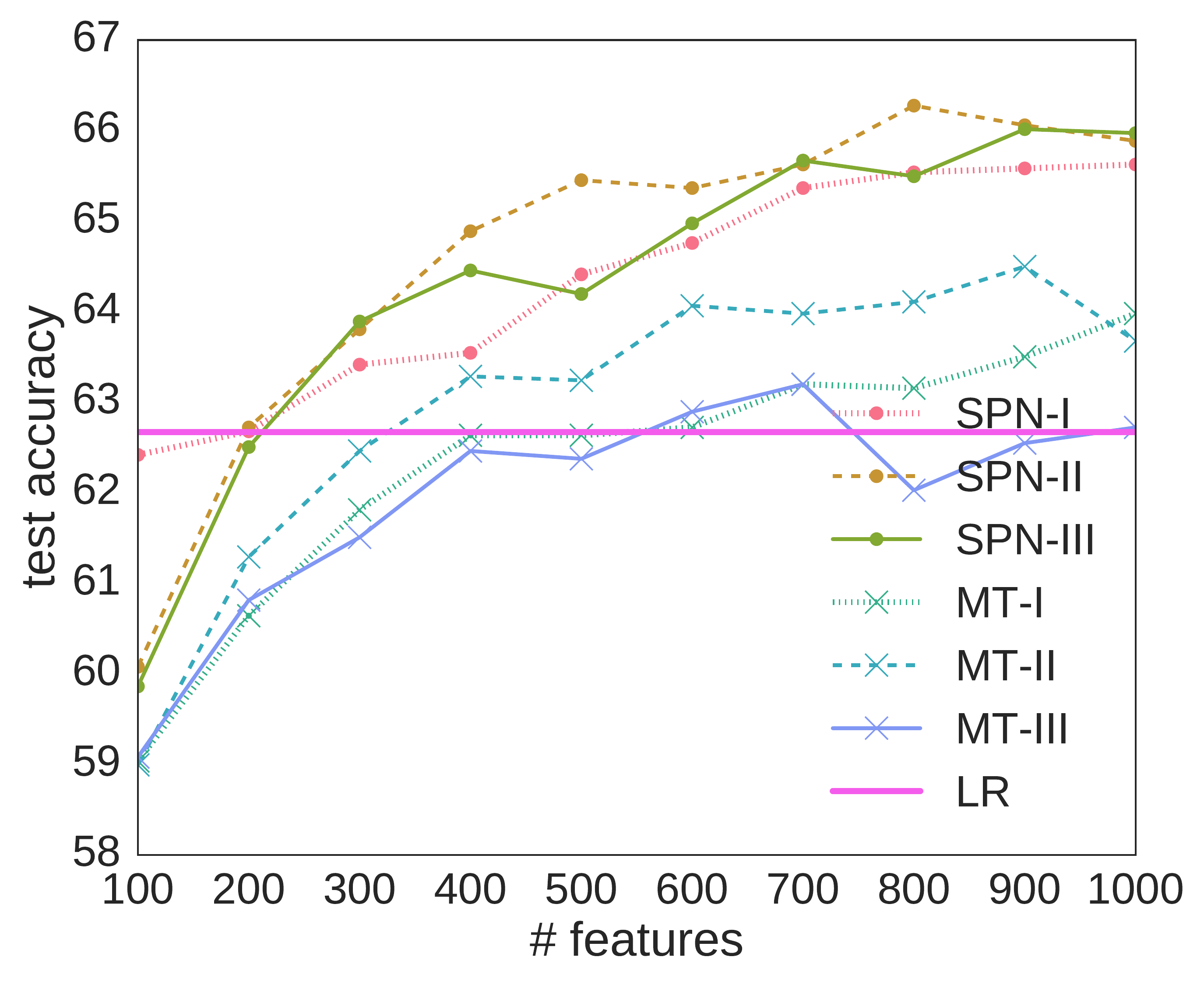}
    \label{fig:rect-lines-cal}}\hspace{-10pt}
  \subfloat[]{\includegraphics[width=0.33\columnwidth]
    {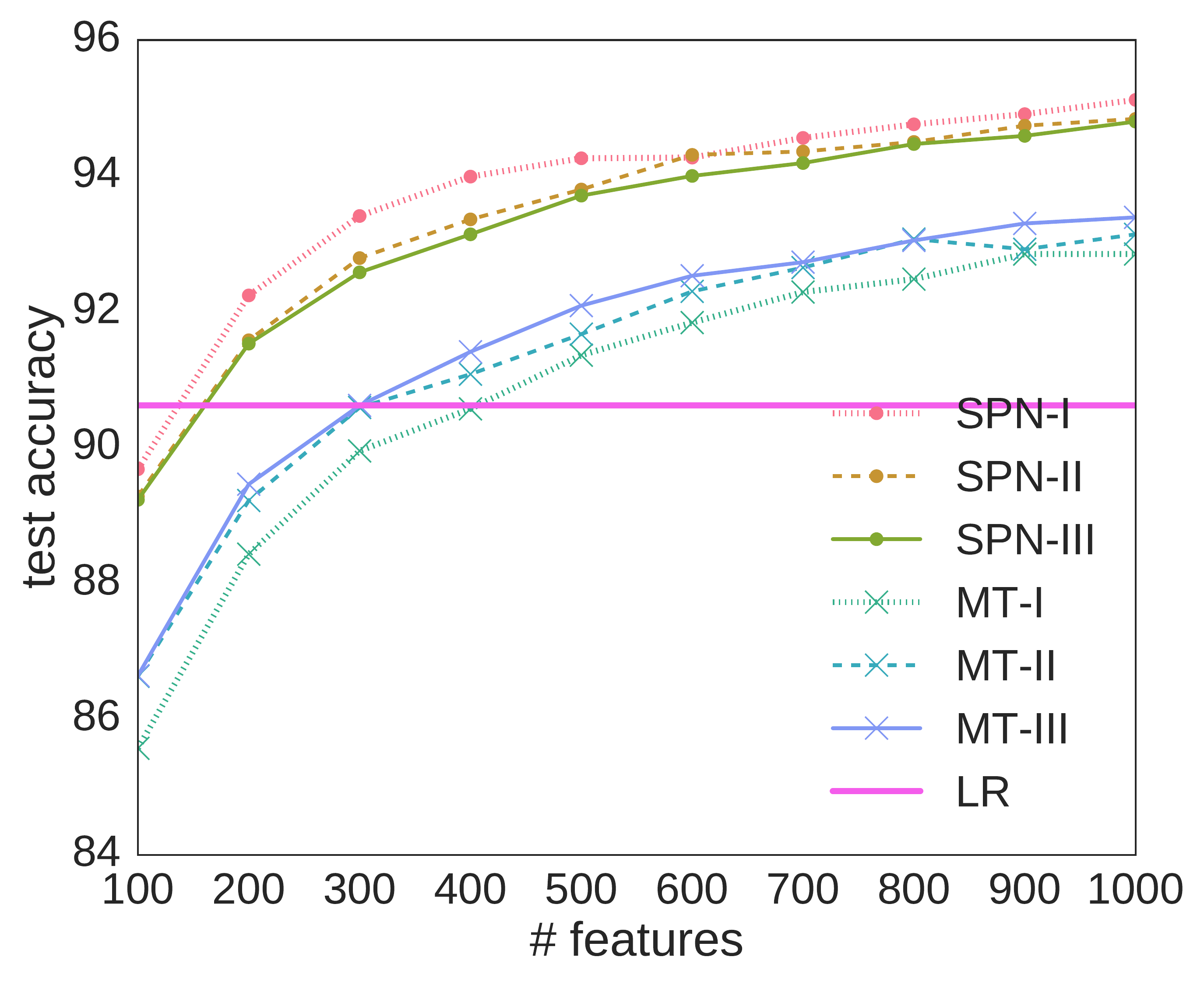}
    \label{fig:rect-lines-bmn}}
  \caption{Test accuracies for \textsf{SPN} and \textsf{MT} models on
    \textsf{REC} (Figs.~\ref{fig:rect-lines-rec}),
    \textsf{CON} (Figs.~\ref{fig:rect-lines-con}),
    \textsf{OCR} (Figs.~\ref{fig:rect-lines-ocr}),
    \textsf{CAL} (Figs.~\ref{fig:rect-lines-cal}) and
    \textsf{BMN} (Figs.~\ref{fig:rect-lines-bmn}).
    against 1000 features generated as random marginal queries
    evaluations.
    Logistic regressor as a baseline is reported as \textsf{LR}.}
  \label{fig:rect-lines}
\end{figure}

Figure~\ref{fig:rect-lines} shows the accuracy scores by \textsf{SPN-I}, \textsf{SPN-II},
\textsf{SPN-III} and \textsf{MT-I}, \textsf{MT-II}, \textsf{MT-III}
on each dataset, while adding 100 features at a time.
The first consideration one can draw is that all models are able to
beat the \textsf{LR} baselines without using all the random features.
On \textsf{OCR}, \textsf{CAL} and \textsf{BMN}, 300 features appear to be enough, for better scoring models, even
200 features provide a nice boost of 1 to 2 points in accuracy
compared to the baselines.
On \textsf{REC} and \textsf{CON}, just 100 features score a very high
accuracy, indicating that the geometric space induced by the
respective models is indeed meaningful.
These considerations hint to the effectiveness of such an approach despite its
randomness.

It is also visible how such embedding accuracies improve as the number of
feature increases.
This is true in almost all scenarios with the exception of
\textsf{REC} and \textsf{CON}.
On these two datasets, a sort of model ``herding'' can be noted: model
performances are not growing noticeably while features are added, with
the exception of the more reguarized model, \textsf{SPN-I} on
\textsf{REC} and of the least regularized one, \textsf{SPN-III} on
\textsf{CON}.
Adding even more features seemed to leave the performance as it is, with a
decreasing pattern when the embedding length has reached 2000. A behavior
that is likely due to the introduction of too many irrelevant features
and to the curse of dimensionality.

Moreover, it is visible how the SPN extracted embeddings outperform the \textsf{MT}
generated ones in almost scenarios, with the exception of
\textsf{CON}.
If we were to compare only the likelihood of these models, the
\textsf{MT} ones would outperform the \textsf{SPN} ones.
One explanation for this behavior can be found in the better capacity
of the SPN structure learner in optimizing the marginal
loglikelihood~\cite{Gens2013}.
Nevertheless, this clearly suggests that a learned representation
comparison can be new informative dimension along which to assess different probabilistic
models.

Concerning the performance of differently regularized models, one can note how more
specialized models like \textsf{SPN-III} and \textsf{SPN-II} tend to perform less well
that \textsf{SPN-I}, a pattern that can be seen for \textsf{MT-III}
on \textsf{CAL} as well.
Such an evidence can be considered as a form of overfitting for the
density estimators.
The greater degree with which they are able to reconstruct and model
the training set, even if does not damage the test data likelihood,
makes the extracted embedding probably too specific, at least for this
set of marginal queries.

\section{Conclusions}
We devised a general and model agnostic schema to extract embeddings
from Tractable Probabilistic Models by exploiting random query
generations and evaluations.
We plan to conduct an extensive empirical experimentation to
investigate in depth several models behavior according to different
embedding evaluation performances.
A proposed empirical comparison on five different standard image
datasets and exploiting random marginal queries hints to the effectiveness of this approach.

All in all, the proposed approaches could be used to compare
differently uncomparable models in a representation learning
framework.
They could serve a role similar to Parzen windows when models
for which the likelihood cannot be easily computed have to be compared
in a generative framework.

More sophisticated query types can improve this general but basic
schema. It would be interesting to correlate the embedding
performances to different query types according to different TPM
models.

\bibliographystyle{splncs03}
\bibliography{referomnia}

\end{document}